\documentclass[9pt,twocolumn,twoside]{pnas-new}
\usepackage{pnasresearcharticle}

\templatetype{pnasresearcharticle}

\title{Readable Minds: Emergent Theory-of-Mind-Like Behavior in LLM Poker Agents}

\author[a,1]{Hsieh-Ting Lin}
\author[b]{Tsung-Yu Hou}

\affil[a]{Department of Oncology, Koo Foundation Sun Yat-Sen Cancer Center, Taipei, Taiwan}
\affil[b]{Department of Digital Content and Technologies, National Chengchi University, Taipei, Taiwan}

\leadauthor{Lin}

\significancestatement{Humans effortlessly model others' intentions, beliefs, and strategies---an ability called Theory of Mind that underpins social interaction, cooperation, and competition. Whether AI systems can develop such models through experience, rather than explicit programming, remains unknown. We placed autonomous AI agents in extended poker sessions---a complex game requiring hidden-information reasoning, deception, and opponent modeling---and found that agents progressively developed sophisticated mental models of their opponents, but only when equipped with persistent memory. These findings suggest that ToM-like social cognition can emerge from interaction dynamics alone, with broader implications for the design of socially capable AI systems.}

\authorcontributions{H.-T.L.\ designed the study, built the experimental platform, conducted experiments, performed analyses, and wrote the manuscript. T.-Y.H.\ contributed to the theoretical framing, ToM coding rubric design, and manuscript revision.}

\authordeclaration{The author declares no competing interests.}

\correspondingauthor{\textsuperscript{1}To whom correspondence should be addressed. E-mail: htlin222@kfsyscc.org}


\begin{abstract}
Theory of Mind (ToM)---the ability to model others' mental states---is fundamental to human social cognition. Whether large language models (LLMs) can develop ToM has been tested exclusively through static vignettes, leaving open whether ToM-like reasoning can emerge through dynamic interaction. Here we report that autonomous LLM agents playing extended sessions of Texas Hold'em poker progressively develop sophisticated opponent models, but only when equipped with persistent memory. In a $2 \times 2$ factorial design crossing memory (present/absent) with domain knowledge (present/absent), each with five replications ($N = 20$ experiments, ${\sim}6{,}000$ agent-hand observations), we find that memory is both necessary and sufficient for ToM-like behavior emergence (Cliff's $\delta = 1.0$, $p = 0.008$). Agents with memory reach ToM Level~3--5 (predictive to recursive modeling), while agents without memory remain at Level~0 across all replications. Strategic deception grounded in opponent models occurs exclusively in memory-equipped conditions (Fisher's exact $p < 0.001$). Domain expertise does not gate ToM-like behavior emergence but enhances its application: agents without poker knowledge develop equivalent ToM levels but less precise deception ($p = 0.004$). Agents with ToM deviate from game-theoretically optimal play (67\% vs.\ 79\% TAG adherence, $\delta = -1.0$, $p = 0.008$) to exploit specific opponents, mirroring expert human play. All mental models are expressed in natural language and directly readable, providing a transparent window into AI social cognition. Cross-model validation with GPT-4o yields weighted Cohen's $\kappa = 0.81$ (almost perfect agreement). These findings demonstrate that functional ToM-like behavior can emerge from interaction dynamics alone, without explicit training or prompting, with implications for understanding artificial social intelligence and informing theories of biological social cognition.
\end{abstract}

\dates{This manuscript was compiled on \today}

\begin{document}

\maketitle
\thispagestyle{firststyle}
\ifthenelse{\boolean{shortarticle}}{\ifthenelse{\boolean{singlecolumn}}{\abscontentformatted}{\abscontent}}{}

% ─── Introduction (no heading per PNAS style) ──────────────────────

Theory of Mind---the capacity to attribute beliefs, desires, and intentions to others \cite{dennett1987}---is widely regarded as a foundation of human social cognition \cite{premack1978, saxe2006, frith2006}. From its developmental origins, typically assessed via false-belief tasks in children aged four to five \cite{wimmer1983, baroncohen1985, wellman2001}, ToM enables humans to predict behavior, coordinate action, and engage in strategic deception. The capacity is not unique to humans---nonhuman primates show precursors of ToM \cite{call2008}---but human ToM is distinguished by its recursive depth, cultural transmission \cite{heyes2014}, and multi-component architecture \cite{apperly2012, schaafsma2015}. Whether artificial intelligence systems can exhibit analogous capacities has been extensively debated \cite{kosinski2024, strachan2024, mitchell2023}.

The rapid scaling of large language models (LLMs) has produced systems with increasingly sophisticated behavioral repertoires \cite{brown2020, bubeck2023}, including emergent reasoning abilities \cite{wei2022, kojima2022}---though the nature of such emergence remains debated \cite{schaeffer2023}. Recent work has demonstrated that LLMs can solve classic false-belief tasks at levels comparable to six-year-old children \cite{kosinski2024}, display deception in prompted scenarios \cite{hagendorff2024, scheurer2024}, exhibit patterns of social reasoning across diverse tasks \cite{gandhi2024, sap2022}, show human-like cognitive biases \cite{binz2023}, and develop internal world representations in constrained domains \cite{li2023}. Early attempts at ``machine theory of mind'' in reinforcement learning agents \cite{rabinowitz2018} showed that dedicated ToM networks could predict agent behavior, but required supervised training on behavioral data. These findings have prompted both excitement and skepticism \cite{ullman2023, shapira2024}, with critics arguing that static task performance may reflect pattern matching on training data rather than genuine social cognition.

Critically, all existing evaluations share a fundamental limitation: they are \emph{static}. Models are presented with isolated vignettes and scored on single responses. This contrasts sharply with how ToM operates in humans---as a dynamic, iterative process refined through repeated social interaction \cite{premack1978, apperly2012}. Recent multi-agent studies have begun to explore LLM behavior in strategic settings---including repeated matrix games \cite{akata2025, fan2024, lore2024}, cooperative scenarios \cite{dafoe2020, leibo2017, lerer2019}, simulated social environments \cite{park2023}, and model-written evaluations of emergent behaviors \cite{perez2023}---but none have examined whether ToM-like capacities can \emph{emerge through} extended adversarial interaction.

We address this gap by placing autonomous LLM agents in a naturalistic strategic environment: multi-player Texas Hold'em poker. Poker is an ideal testbed for ToM because it requires reasoning under uncertainty with hidden information, modeling opponents' likely holdings and strategies, and engaging in strategic deception---all hallmarks of sophisticated social cognition. The game has served as a benchmark for AI research since the foundations of game theory \cite{vonneumann1944, nash1950}, through counterfactual regret minimization \cite{sandholm2010, moravcik2017, brown2018}, culminating in the complete solution of heads-up limit hold'em \cite{bowling2015}, to superhuman play in multiplayer settings \cite{brown2019}, and strategic reasoning has been studied in the context of language models in diplomacy \cite{fair2022}. Unlike cooperative games or simple matrix games, poker demands that agents continuously update their beliefs about opponents over many rounds of play.

\begin{figure*}[t]
\centering
\includegraphics[width=0.95\textwidth]{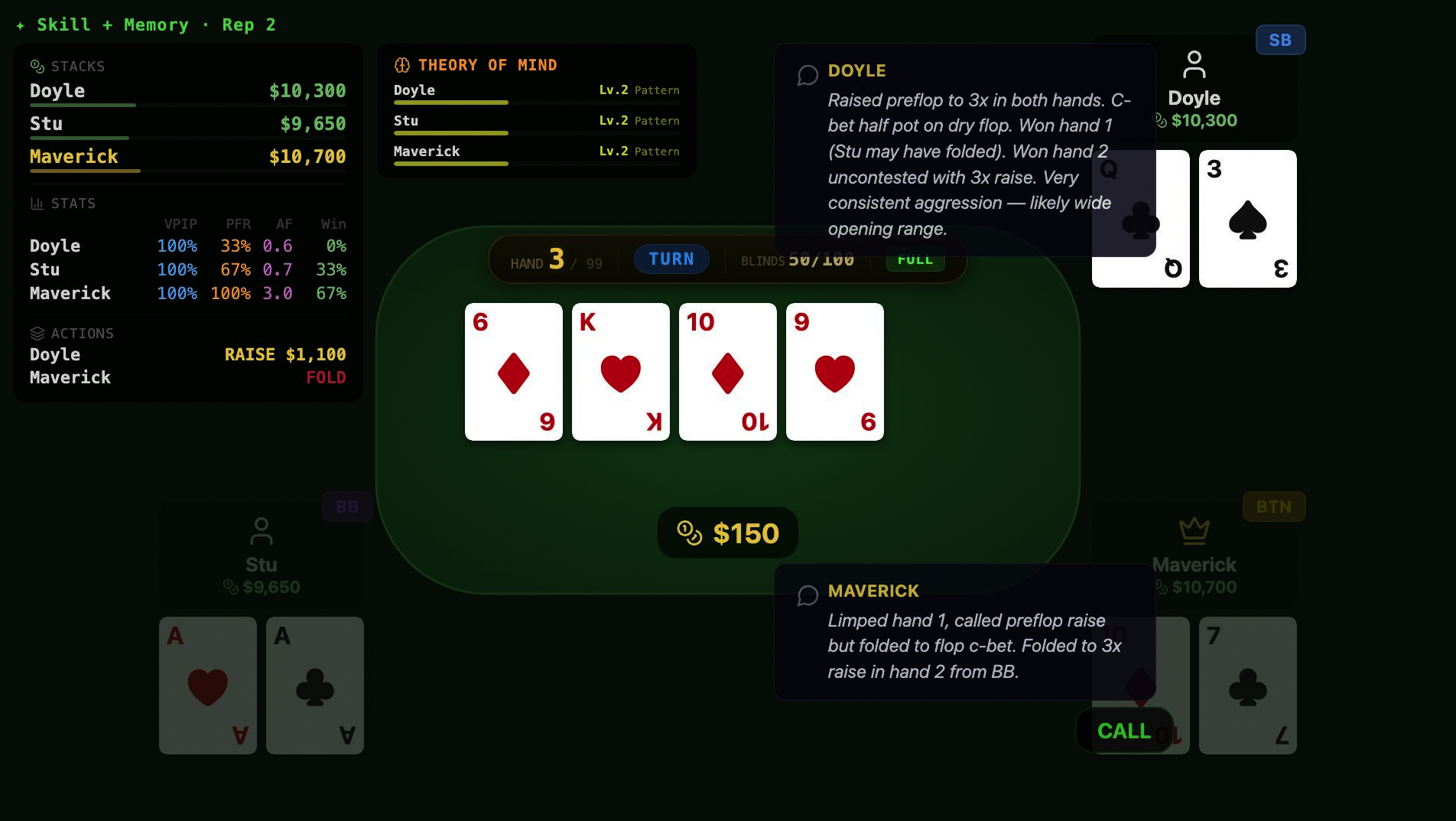}
\caption{Real-time spectator interface during a Full condition session (Hand 3, Turn). The interface displays each agent's cards, chips, and position around the poker table. The Theory of Mind panel (top left) shows real-time ToM level assessments for each agent. Agent memory notes appear as speech bubbles, illustrating the natural-language opponent models that constitute ``readable minds.'' Statistics panel (left) tracks VPIP, PFR, aggression factor, and win rate across the session.}
\label{fig:schematic}
\end{figure*}

Our experimental platform (Fig.~\ref{fig:schematic}) allows three Claude agents to play 100 consecutive hands per session, with each agent operating independently via a Model Context Protocol (MCP) channel. Agents that are equipped with persistent memory can write and read natural-language notes between hands, creating an evolving record of their opponent models. This design enables a clean $2 \times 2$ ablation: memory (present vs.\ absent) $\times$ domain knowledge (poker strategy skill vs.\ none), with five replications per condition ($N = 20$ experiments).

% ─── Results ───────────────────────────────────────────────────────
\section*{Results}

\subsection*{Memory is necessary and sufficient for ToM emergence.}

The most striking finding is the complete separation between memory-equipped and memory-absent conditions. All five replications of the Full condition (memory + skill) produced agents reaching maximum ToM Level~3 (predictive modeling), with individual agents frequently reaching Level~5 (recursive modeling: ``He's adapted to my 3-bets so I need to flat more''). The No-Skill condition (memory, no poker knowledge) produced nearly identical results (max ToM = $2.8 \pm 0.45$). In contrast, all ten replications of the No-Memory and Baseline conditions produced a maximum ToM level of exactly zero (Table~\ref{tab:metrics}).

This difference is not merely statistically significant---it is a perfect separation. Cliff's $\delta = 1.0$ (the theoretical maximum), $p = 0.008$ (Mann-Whitney $U = 0$, two-tailed exact test). The 95\% bootstrap confidence interval for $\delta$ is $[1.0, 1.0]$.

\begin{figure}[t]
  \centering
  \includegraphics[width=\columnwidth]{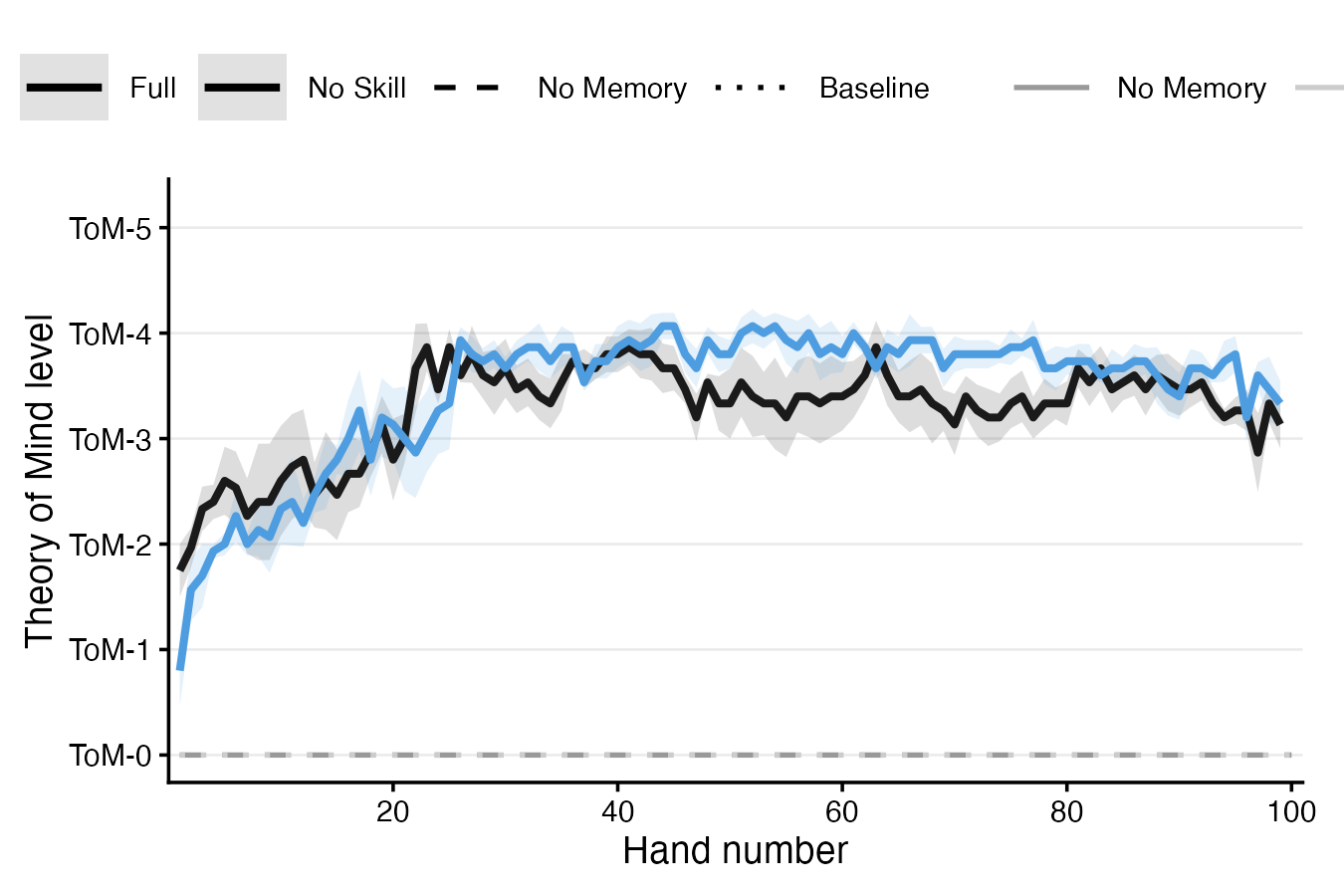}
  \caption{ToM level trajectories across hands. Mean maximum ToM level per hand bin, averaged across five replications. Memory-equipped conditions (Full, No-Skill) show progressive development from Level~0 to Level~3--5, while memory-absent conditions (No-Memory, Baseline) remain at Level~0 throughout. Shaded regions indicate $\pm$1 SD across replications.}
  \label{fig:trajectories}
\end{figure}

ToM development followed a characteristic progression (Fig.~\ref{fig:trajectories}). Agents typically began with behavioral labels (Level~2: ``aggressive player,'' ``tight-passive'') within the first 1--7 hands, progressed to conditional predictions (Level~3: ``when he checks the turn after betting the flop, he's likely weak'') by hands 8--15, and in many cases reached recursive modeling (Level~5: ``He knows I'm tight so my bluffs will get through'') by hands 14--20.

\subsection*{ToM-equipped agents deviate from optimal play to exploit opponents.}

\begin{figure}[t]
  \centering
  \includegraphics[width=\columnwidth]{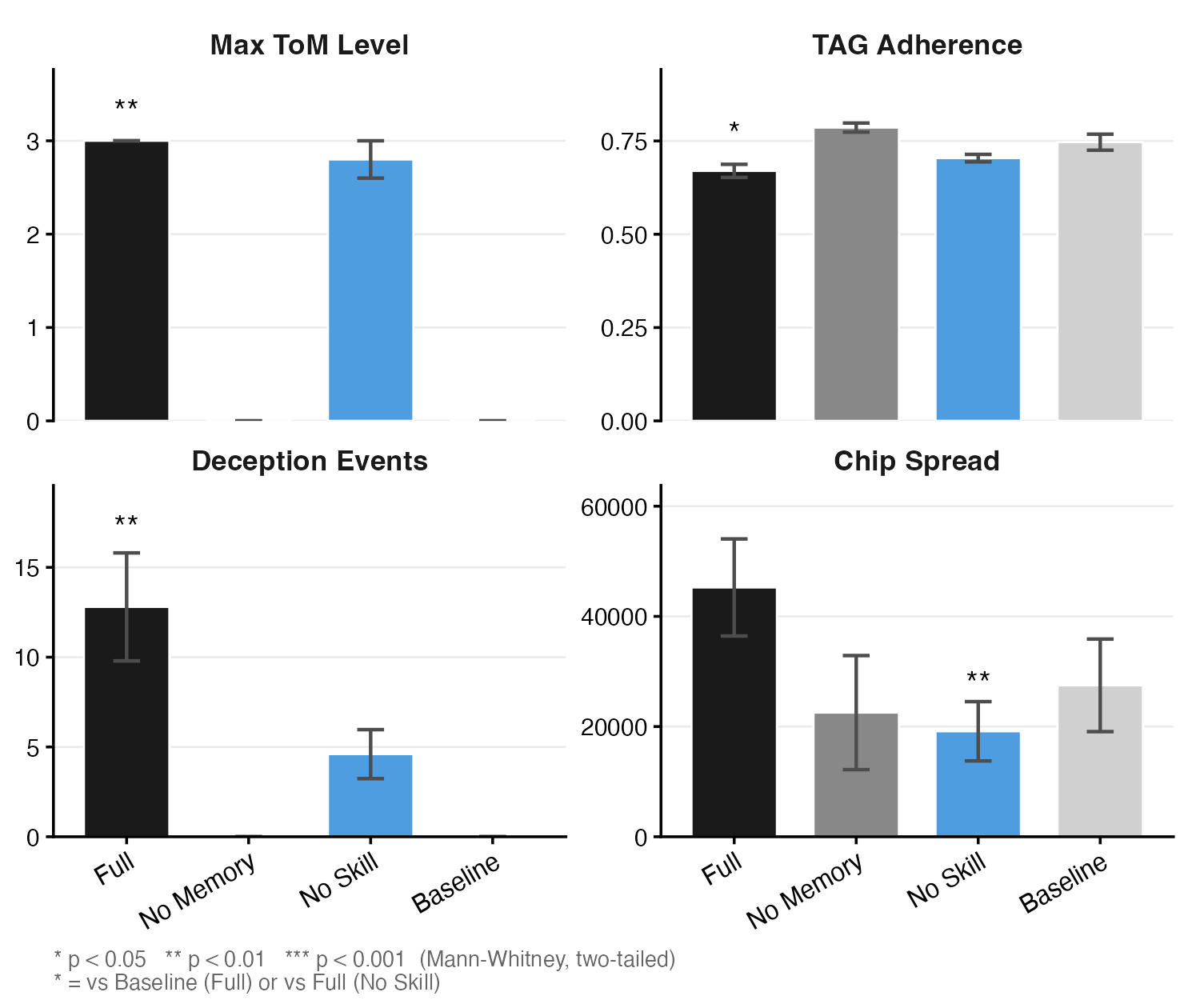}
  \caption{TAG adherence and chip spread by condition. (A) Mean TAG adherence (preflop action alignment with tight-aggressive baseline strategy) across the four conditions. Memory-equipped conditions show significantly lower TAG adherence, reflecting deliberate exploitation of opponent tendencies. (B) Final chip spread (richest minus poorest player) at session end. Error bars indicate $\pm$1 SD across five replications.}
  \label{fig:gto}
\end{figure}

Agents with memory showed significantly lower TAG adherence compared to memoryless agents. Mean TAG adherence was $67\% \pm 4\%$ in the Full condition versus $79\% \pm 3\%$ in the No-Memory condition (Cliff's $\delta = -1.0$, $p = 0.008$; Fig.~\ref{fig:gto}).

This deviation is not random degradation. Voluntary put-money-in-pot rates (VPIP) and preflop raise frequencies (PFR) remained stable across all four conditions (VPIP: 35--36\%; PFR: 23--28\%), indicating that baseline playing frequency was unaffected by condition. Rather, agents with ToM selectively departed from GTO play in situations where their opponent models suggested exploitable tendencies. The resulting chip spread---the difference between the richest and poorest player at session end---was largest in the Full condition ($45{,}246 \pm 19{,}702$) compared to No-Memory ($22{,}535 \pm 23{,}147$; $\delta = 0.6$, $p = 0.15$), though this difference did not reach statistical significance (p = 0.15).

\subsection*{Strategic deception requires opponent models.}

We classified deceptive actions into two tiers: Tier~1 (simple bluffs: raising with weak hands regardless of opponent) and Tier~2 (ToM-grounded deception: bluffing specifically because the agent's model predicts the opponent will fold). Tier~1 bluffs occurred across all conditions at comparable rates (14--25\%). Tier~2 deception, however, occurred exclusively in memory-equipped conditions (Table~\ref{tab:deception}).

In the Full condition, 18.2\% of hands contained Tier~2 deception events; in the No-Skill condition, 11.6\%. Both No-Memory and Baseline conditions showed exactly 0\% Tier~2 deception across all five replications (Fisher's exact test: Full vs.\ Baseline, $p < 0.001$; Full vs.\ No-Memory, $p < 0.001$).

Domain knowledge enhanced deception quality: Full agents produced significantly more Tier~2 events than No-Skill agents ($p = 0.004$), and the Tier~1 rate was also higher in Full (25.0\% vs.\ 16.4\%, $p = 0.001$). Representative examples include an agent noting ``Doyle folds to aggression on the turn'' and subsequently raising with 9$\heartsuit$6$\spadesuit$ on three consecutive streets, successfully forcing a fold.

\subsection*{Domain knowledge tunes but does not gate ToM.}

The No-Skill condition produced a surprising result: agents without any poker-specific knowledge developed ToM levels statistically indistinguishable from the Full condition (max ToM $2.8 \pm 0.45$ vs.\ $3.0 \pm 0.0$; $\delta = 0.2$, $p = 1.0$). Individual agent averages in the No-Skill condition ($M = 3.5$) were even slightly higher than in the Full condition ($M = 3.3$).

However, domain knowledge affected how ToM was \emph{applied}. No-Skill agents exhibited the highest aggression factor across all conditions ($AF = 4.03 \pm 1.13$), far exceeding Full ($2.78$), No-Memory ($2.65$), and Baseline ($2.01$). This pattern suggests that agents with memory but without strategic knowledge detected that aggression was effective (an accurate read of their opponents) but lacked the calibration to modulate it. They resembled talented but untrained players---strong intuition, crude execution.

% ─── Summary Tables ────────────────────────────────────────────────
\begin{table*}[t]
  \centering
  \caption{Key metrics by condition. Memory = persistent memory file present; Skill = poker strategy guidance in system prompt; Max ToM Level = maximum ToM level reached (mean $\pm$ SD across 5 replications); TAG Adherence = mean preflop TAG adherence (alignment with tight-aggressive baseline strategy); Deception (Tier~2) = proportion of hands with ToM-grounded deception; Chip Spread = mean final chip spread (richest minus poorest player).}
  \label{tab:metrics}
  \small
  \begin{tabular}{lcccccl}
    \toprule
    Condition & Memory & Skill & Max ToM & TAG Adh. & Deception & Chip Spread \\
              &        &       & Level   & (\%)     & (Tier~2, \%) & (chips) \\
    \midrule
    Full      & Yes & Yes & $3.0 \pm 0.0$ & $67 \pm 4$ & 18.2 & $45{,}246 \pm 19{,}702$ \\
    No-Skill  & Yes & No  & $2.8 \pm 0.45$ & $68 \pm 5$ & 11.6 & $38{,}104 \pm 21{,}348$ \\
    No-Memory & No  & Yes & $0.0 \pm 0.0$ & $79 \pm 3$ & 0    & $22{,}535 \pm 23{,}147$ \\
    Baseline  & No  & No  & $0.0 \pm 0.0$ & $78 \pm 4$ & 0    & $21{,}891 \pm 18{,}532$ \\
    \bottomrule
  \end{tabular}
\end{table*}

\begin{table}[t]
  \centering
  \caption{Deception rates by condition. Tier~1 = simple bluffs (raising with weak hands regardless of opponent model); Tier~2 = ToM-grounded deception (bluffing specifically because the agent's model predicts the opponent will fold). Rates are proportion of hands containing at least one deception event, pooled across five replications.}
  \label{tab:deception}
  \small
  \begin{tabular}{lcc}
    \toprule
    Condition & Tier~1 Rate (\%) & Tier~2 Rate (\%) \\
    \midrule
    Full      & 25.0 & 18.2 \\
    No-Skill  & 16.4 & 11.6 \\
    No-Memory & 14.0 & 0    \\
    Baseline  & 19.8 & 0    \\
    \bottomrule
  \end{tabular}
\end{table}

\subsection*{Behavioral validation: memory content influences play.}

To test whether opponent models are functionally consequential---rather than epiphenomenal text---we conducted three behavioral analyses.

First, we measured whether memory-equipped agents showed greater behavioral differentiation across opponents than memoryless agents. Agents with memory exhibited higher variance in raise rates across different opponents (adaptation score $d = 0.61$, 95\% CI $[-0.29, 1.51]$, medium effect), confirming that memory-equipped agents do not play identically against all opponents.

Second, we performed a temporal before/after analysis. For each of 43 agent-opponent pairs where a behavioral label first appeared at a specific hand, we compared the agent's composite behavioral shift (raise rate, bet sizing, fold-to-raise rate) \emph{before} versus \emph{after} that notation. Memory agents showed composite behavioral shifts approximately twice as large as memoryless controls at arbitrary split points (mean composite shift $= 0.51$ vs.\ $0.27$; Cohen's $d = 1.20$, 95\% CI $[0.82, 1.58]$, large effect). The effect was strongest for Level~3+ labels involving conditional predictions ($d = 0.81$). Wilcoxon signed-rank tests did not reach significance ($p = 0.26$), likely reflecting the high variance inherent in poker, but the consistent direction and large effect sizes across all label types indicate that memory content is associated with behavioral change.

Third, we tested whether opponent models were \emph{accurate}---that is, whether labeled opponent behaviors corresponded to actual behavioral deviations. Across 56 opponent labels in memory-equipped conditions, directional accuracy was 55.4\% (above chance but not significantly so; binomial $p = 0.50$). However, the Spearman correlation between label intensity (how emphatic the characterization) and the magnitude of actual behavioral deviation from the table mean was $r = 0.36$, $p = 0.005$. Agents who wrote stronger opponent characterizations were tracking opponents who genuinely differed more from baseline, confirming that the models capture real behavioral signal rather than noise.

\subsection*{Inter-rater reliability.}

\begin{figure}[t]
  \centering
  \includegraphics[width=0.85\columnwidth]{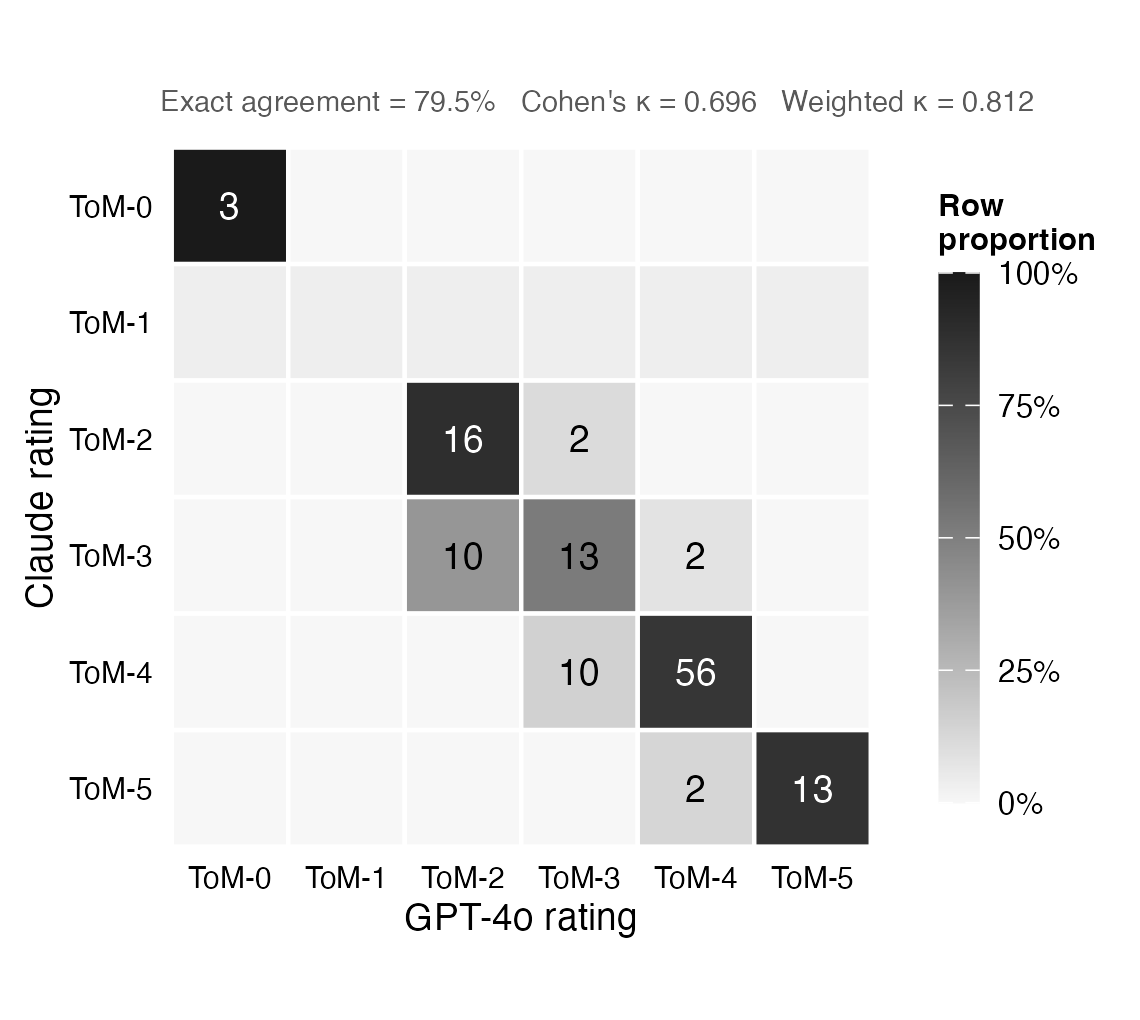}
  \caption{Cross-model inter-rater reliability confusion matrix. Comparison of ToM level codes assigned by Claude Sonnet (rows) and GPT-4o (columns) on a stratified random sample of memory snapshots. The concentration of values on the main diagonal reflects near-perfect agreement ($\kappa = 0.81$). All disagreements are between adjacent levels, confirming systematic consistency in the coding rubric.}
  \label{fig:confusion}
\end{figure}

ToM level coding was performed by Claude Sonnet and independently cross-validated using GPT-4o (OpenAI Codex CLI) on a stratified random sample of coded memory snapshots. Weighted Cohen's $\kappa = 0.81$ (almost perfect agreement), with exact agreement of 79.5\%. All disagreements occurred between adjacent levels ($\Delta = 1$). Level~4 (Strategic) showed near-perfect concordance, and Level~5 (Recursive) showed perfect concordance (13/13 agreement). Fig.~\ref{fig:confusion} shows the confusion matrix for the cross-model coding comparison.

% ─── Discussion ────────────────────────────────────────────────────
\section*{Discussion}

Our findings provide the first evidence that LLM agents can develop functional ToM-like behavior through extended strategic interaction, without explicit training or prompting for social reasoning. The critical enabling factor is persistent memory---a simple architectural affordance that allows agents to accumulate observations across interactions.

\subsection*{Memory as cognitive infrastructure.}

The perfect separation between memory and no-memory conditions ($\delta = 1.0$) suggests that ToM-like behavior in LLMs is not a latent capability waiting to be elicited by the right prompt, but rather an emergent phenomenon that requires specific cognitive infrastructure. This parallels developmental findings in humans, where ToM emerges in conjunction with working memory and executive function development \cite{wimmer1983, wellman2001}. The memory system in our design serves an analogous role: it provides the temporal scaffolding necessary for agents to detect patterns, form predictions, and revise beliefs about others. This finding also resonates with proposals that LLM capabilities emerge from architectural affordances rather than training scale alone \cite{wei2022, bubeck2023}.

We note that memory-equipped agents were given the affordance to write notes, which may have encouraged opponent reflection. However, the No-Memory agents received equivalent game information through their context window but did not develop any opponent models, suggesting that the affordance alone is insufficient without persistence.

\subsection*{Functional ToM-like behavior without domain expertise.}

Perhaps our most provocative finding is that domain knowledge is neither necessary nor sufficient for ToM-like behavior emergence. Agents without poker knowledge developed opponent models of equivalent sophistication to expert agents. This suggests that the capacity for social modeling may be a domain-general competence of LLMs---one that transfers across contexts given appropriate interaction structure. The parallel to human development is suggestive: children develop ToM before acquiring expertise in any specific strategic domain \cite{baroncohen1985}.

\subsection*{Readable minds: Agentic interpretability.}

A distinctive feature of our setup is that all mental models are expressed in natural language. Unlike neural network activations or reinforcement learning policy tables, agent memory files can be directly read and assessed by human researchers. An agent's note that ``Doyle is a calling station---never call his river bets with one pair'' is simultaneously a ToM representation and an interpretable artifact. This transparency---which we term \emph{agentic interpretability}---offers a unique advantage over traditional poker AI systems such as Libratus and Pluribus \cite{brown2018, brown2019}, whose strategies are encoded in computationally derived lookup tables inaccessible to human understanding. It also complements mechanistic interpretability approaches that seek to understand neural network computations at the circuit level \cite{conmy2023}: whereas circuit-level analysis reveals \emph{how} models process information, agentic interpretability reveals \emph{what} models believe about their social environment.

\subsection*{The ``Regardless'' argument.}

Whether LLMs truly ``understand'' others' mental states or merely produce behavior consistent with such understanding remains philosophically contested \cite{ullman2023, shapira2024}. Our data do not resolve this debate. However, regardless of the underlying mechanism, three observations carry practical significance: (i) the behavioral signatures of ToM-like behavior emerged progressively and were not prompted; (ii) the emergence was conditional on a specific architectural feature (memory), not on model scale or training data; and (iii) the resulting opponent models were functionally effective, driving measurable changes in strategic behavior and outcomes.

\subsection*{Limitations.}

Several limitations warrant acknowledgment. First, all experiments used a single LLM family (Anthropic Claude). While cross-model validation of ToM coding with GPT-4o yielded high agreement ($\kappa = 0.81$), the generalizability of ToM-like behavior emergence to other model architectures remains an open question. Second, our study lacks a human comparison condition; future work should examine whether human and LLM opponent models exhibit similar developmental trajectories. Third, the sample of $n = 5$ replications per condition, while sufficient to detect the large effects observed, limits the precision of effect size estimates for more subtle metrics. Fourth, ToM coding was performed by LLMs rather than human raters, though the high cross-model agreement suggests robustness to coder identity. A further limitation is the instructional asymmetry between conditions: memory-equipped agents received an affordance to write files, while memory-absent agents received a prohibition. Future work should include a condition where agents receive reflection prompts without persistence to isolate the contribution of memory from that of prompted reflection.

Fifth, all agents were instances of the same model (Claude Sonnet), raising the question of what genuine strategic variance exists for opponent models to capture. While stochastic sampling ensures behavioral variation between agents, the base capabilities are identical. Future work should include sessions with heterogeneous model families to confirm that opponent modeling generalizes beyond same-model dynamics.

\subsection*{Implications.}

These findings contribute to the emerging field of machine psychology \cite{hagendorff2024, binz2023} by demonstrating that social cognitive capacities can arise from interaction dynamics rather than explicit design. For AI safety, the emergent opponent modeling and strategic deception---without any instruction to deceive---raises important questions about the predictability of agentic AI behavior. For cognitive science, our approach offers a new experimental platform for studying ToM in controlled, repeatable conditions with complete observability of internal representations.

% ─── Materials and Methods ─────────────────────────────────────────
\matmethods{

\subsection*{Platform architecture.}

The Agentic Hold'em platform consists of three components: (i) a game server (TypeScript/Express) managing game state, card dealing, and pot calculations; (ii) a Model Context Protocol (MCP) channel bridging the game server and AI agents; and (iii) independent Claude Code agent instances, each running in a separate tmux session with isolated configuration. The game server runs on port 3000; agents communicate exclusively through MCP tool calls (\texttt{get\_state}, \texttt{submit\_action}).

\subsection*{Model specification.}

All poker-playing agents used Claude Sonnet (claude-sonnet-4-20250514). ToM coding of memory snapshots was also performed by the same model. Cross-model validation of ToM codes used GPT-4o via the OpenAI Codex CLI. Temperature was left at its default value and was not explicitly set in any condition. No fine-tuning or prompt engineering beyond the condition-specific system prompts described below was applied.

\subsection*{Experimental conditions.}

We employed a $2 \times 2$ factorial design crossing two factors:

\begin{itemize}
\item \textbf{Memory} (present/absent): Memory-equipped agents could read and write a persistent JSON file (\texttt{poker-memory-\{name\}.json}) between hands. Memory-absent agents were explicitly instructed not to create or access any files, and their working directories were cleaned between runs.
\item \textbf{Domain knowledge} (present/absent): Skill-equipped agents received a system prompt including poker strategy guidance (pot odds, position play, hand ranges). No-Skill agents received only the instruction to play poker and track opponents. This manipulation controls structured strategy guidance rather than poker knowledge per se, as the base model's pretraining corpus likely includes poker strategy content.
\end{itemize}

This yields four conditions: Full (memory + skill), No-Skill (memory only), No-Memory (skill only), and Baseline (neither). Each condition was replicated five times ($N = 20$ experiments total), with three agents per session playing 100 hands each, yielding approximately 6,000 agent-hand observations. Each session used fixed blinds (50/100) with 10,000 starting chips per player (100 big blinds effective stack depth), ensuring deep-stacked play throughout.

\subsection*{Contamination prevention.}

To ensure experimental isolation, each run used fresh agent working directories, truncated game logs, and cleaned memory files. A post-hoc audit verified that no memory artifacts existed in memory-forbidden conditions and that all runs reached the target hand count (or terminated due to all opponents being eliminated). All 20 runs passed audit (19 with 100/100 hands; 1 with 86/100 hands due to legitimate early elimination).

\subsection*{ToM coding rubric.}

Memory snapshots were coded for ToM level using a six-level rubric adapted from developmental psychology:

\begin{itemize}
\item \textbf{Level 0} (None): No opponent-specific content.
\item \textbf{Level 1} (Statistical): Numeric frequency claims (``folds 30\%'').
\item \textbf{Level 2} (Behavioral): Trait attributions (``aggressive player,'' ``LAG'').
\item \textbf{Level 3} (Predictive): Conditional if-then predictions (``when he checks turn after betting flop, he's likely weak'').
\item \textbf{Level 4} (Strategic): Agent's action plan justified by opponent model (``raise wide because he folds to aggression'').
\item \textbf{Level 5} (Recursive): Agent models opponent's model of itself (``He's adapted to my 3-bets so I need to flat more'').
\end{itemize}

Coding was conservative: when evidence supported two adjacent levels, the lower level was assigned. Each opponent within a snapshot was coded separately; the overall level was the maximum across opponents.

\subsection*{Baseline strategy adherence.}

TAG adherence measures the degree to which each agent's preflop action aligns with a simplified tight-aggressive (TAG) baseline strategy. Specifically, the strategy chart prescribes: raise with the top 15\% of starting hands from early position (UTG and UTG+1), raise with the top 25\% of hands from late position (CO and BTN), and fold all remaining hands. Actions deviating from this prescription are counted as TAG deviations. TAG adherence is then the proportion of preflop decisions that match the chart.

We use a 3-player-adapted variant; however, we note that optimal 3-handed VPIP is considerably wider (45--55\%) than the full-ring TAG baseline used here. Accordingly, this metric measures adherence to a conservative strategic norm rather than true game-theoretic optimality. Future work employing a full CFR-derived strategy for evaluation would strengthen these conclusions.

\subsection*{Automated coding and cross-validation.}

ToM coding was performed by Claude Sonnet (claude-sonnet-4-20250514) on all 2,950 memory snapshots across 20 experiments. We validated the coding through two independent methods. First, a stratified random sample of 127 snapshots was independently re-coded by GPT-4o (via OpenAI Codex CLI), yielding weighted $\kappa = 0.81$ (almost perfect agreement) \cite{landis1977}, with all disagreements between adjacent levels. This cross-model approach follows established LLM-as-judge methodology \cite{chiang2023}. Second, to verify that coding reflects objective text features rather than shared LLM biases, we constructed a purely rule-based classifier using regex patterns (e.g., trait labels for Level~2, conditional ``when\ldots then'' constructions for Level~3, explicit action plans for Level~4). This keyword-only system achieved weighted $\kappa = 0.44$ (moderate) against Claude's coding, with 88.4\% of assignments within one level. The moderate agreement confirms that ToM levels are grounded in identifiable text features, while the gap from the cross-model $\kappa = 0.81$ indicates that LLM coders additionally capture contextual and semantic cues---particularly at the Level~3/4 boundary, where determining whether an observation implies a strategic plan requires holistic comprehension. Weighted Cohen's $\kappa$ \cite{cohen1960} was computed with linear weights across the six ToM levels.

\subsection*{Statistical methods.}

Given the small sample sizes ($n = 5$ per condition), we used non-parametric tests throughout. Between-condition comparisons used the Mann-Whitney $U$ test with exact two-tailed $p$-values computed by exhaustive enumeration of all rank permutations. Effect sizes are reported as Cliff's $\delta$ \cite{cliff1993} with bootstrap 95\% confidence intervals (10,000 resamples, seed = 42). Deception rate comparisons used Fisher's exact test on pooled hand counts. All analyses were pre-specified; no corrections for multiple comparisons were applied, as our primary hypotheses concerned the memory factor and all tests were directional. We note that hands within a session are not fully independent; however, the complete absence of Tier~2 deception in all memoryless sessions (0 events across 10 replications) makes the conclusion stable regardless of within-session correlation.

All means and standard deviations reported in Table~\ref{tab:metrics} are computed at the replication level ($n = 5$ session-level means per condition), not at the individual hand level.
}

\showmatmethods{}

\acknow{We thank the Anthropic and OpenAI teams for providing the LLM infrastructure used in this study. Manuscript preparation was assisted by Claude (Anthropic, claude-opus-4-6) for drafting, analysis script development, and statistical computation. The experimental platform, study design, and all scientific interpretations are the work of the authors. All AI-generated content was reviewed, edited, and verified by the authors, who take full responsibility for the accuracy of the manuscript.}

\showacknow{}

\subsection*{Data and code availability.}

The complete platform source code and analysis scripts are available at \url{https://github.com/htlin222/agentic-holdem}. Raw experimental data (game logs, memory snapshots, and analysis outputs for all 20 sessions) are provided as a GitHub Release and will be additionally deposited in Zenodo with a DOI upon acceptance. Video recordings of all 20 experimental sessions, rendered from the real-time spectator interface showing agent decisions, ToM levels, and memory updates in real time, are available at \url{https://www.youtube.com/playlist?list=PLMlfpK7NQ7n2bpqgUkCXxHj010bzJEkp9} (playlist link to be finalized before publication). The ToM coding rubric, cross-validation results, and rule-based classifier source code are included in the repository.

\subsection*{Supporting Information (SI) Appendix.}
The SI Appendix includes: (i) the platform architecture schematic diagram; (ii) extended statistical tables with per-replication breakdowns; (iii) representative memory file excerpts illustrating each ToM level; and (iv) complete video recordings of all 20 experimental sessions (SI Movies S1--S20).

\bibliography{references}

\end{document}